\pdfoutput=1
\documentclass[11pt]{article}

\usepackage[]{ACL2023}

\usepackage{times}
\usepackage{latexsym}
\usepackage{graphicx}
\usepackage{booktabs}
\usepackage{pifont}
\usepackage{amsmath}
\usepackage{color}
\usepackage{bm}
\usepackage{amsfonts}
\usepackage{multirow}

\usepackage[T1]{fontenc}
\usepackage[utf8]{inputenc}
\newcommand{\mymodel}{PlugD}
\newcommand{\setting}{plug-and-play representation learning}

\usepackage{microtype}

\usepackage{inconsolata}

\title{Plug-and-Play Document Modules for Pre-trained Models}

\author{
Chaojun Xiao\textsuperscript{\rm 1,2,3}, 
Zhengyan Zhang\textsuperscript{\rm 1,2,3},
Xu Han\textsuperscript{\rm 1,2,3}\thanks{~~Corresponding authors.}~,
Chi-Min Chan\textsuperscript{\rm 1,2,3},
Yankai Lin\textsuperscript{\rm 4,5}\\
\textbf{
Zhiyuan Liu\textsuperscript{\rm 1,2,3}{$^*$},
Xiangyang Li\textsuperscript{\rm 6},
Zhonghua Li\textsuperscript{\rm 6}, 
Zhao Cao\textsuperscript{\rm 6},
Maosong Sun\textsuperscript{\rm 1,2,3}{$^*$}} \\
\textsuperscript{\rm 1}NLP Group, DCST, IAI, BNRIST, Tsinghua University, Beijing \\
\textsuperscript{\rm 2}International Innovation Center of Tsinghua University, Shanghai \textsuperscript{\rm 3}Quan Cheng Laboratory\\
\textsuperscript{\rm 4}Gaoling School of Artificial Intelligence, Renmin University of China, Beijing \\
\textsuperscript{\rm 5}Beijing Key Laboratory of Big Data Management and Analysis Methods \\
\textsuperscript{\rm 6}Huawei Technologies Co., Ltd. \\
\texttt{xiaocj20@mails.tsinghua.edu.cn}, \texttt{\{hanxu2022,liuzy,sms\}@tsinghua.edu.cn}}

\begin{document}

\maketitle
\begin{abstract}
Large-scale pre-trained models (PTMs) have been widely used in document-oriented NLP tasks, such as question answering. However, the encoding-task coupling requirement results in the repeated encoding of the same documents for different tasks and queries, which is highly computationally inefficient. To this end, we target to decouple document encoding from downstream tasks, and propose to represent each document as a plug-and-play document module, i.e., a document plugin, for PTMs (\mymodel{}). By inserting document plugins into the backbone PTM for downstream tasks, we can encode a document one time to handle multiple tasks, which is more efficient than conventional encoding-task coupling methods that simultaneously encode documents and input queries using task-specific encoders. Extensive experiments on 8 datasets of 4 typical NLP tasks show that \mymodel{} enables models to encode documents once and for all across different scenarios. Especially, \mymodel{} can save $69\%$ computational costs while achieving comparable performance to state-of-the-art encoding-task coupling methods. Additionally, we show that \mymodel{} can serve as an effective post-processing way to inject knowledge into task-specific models, improving model performance without any additional model training. Our code and checkpoints can be found in \url{https://github.com/thunlp/Document-Plugin}.

\end{abstract}

\section{Introduction}

In recent years, large-scale pre-trained models (PTMs)~\cite{BERT,t5} have been widely adopted and achieved breakthrough performance for document-oriented NLP tasks, such as question answering. However, due to the tight coupling of document encoding and concrete tasks, PTMs have to dynamically generate document representations according to specific tasks and queries, leading to the repeated encoding of the same documents in different applications. For example, Wikipedia documents are commonly used in various knowledge-intensive tasks such as question answering~\cite{openqa-drqa}, fact verification~\cite{FEVER}, and dialogue generation~\cite{WoW}. 
In this case, existing methods separately encode one document for each task or even for each input query (e.g., a question for question answering, a claim for fact verification), making them highly computationally inefficient. 
To this end, it raises a natural question: \textit{can we decouple document encoding from concrete tasks, encoding documents only once and with guaranteed transferability across multiple tasks?}

\begin{figure}[t]
    \centering
    \includegraphics[width=\columnwidth]{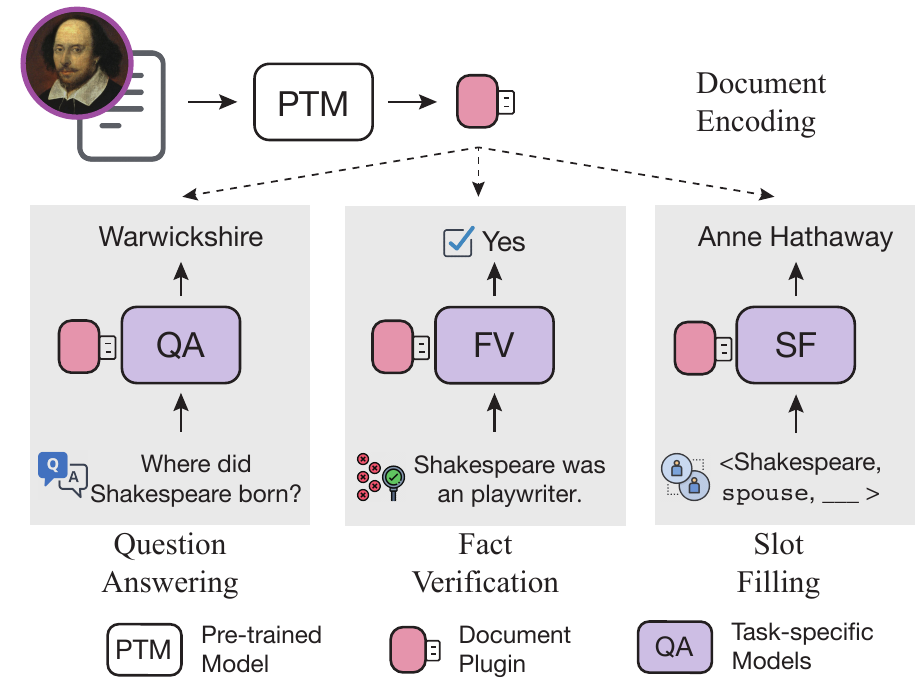}
    \caption{Illustration of plug-and-play document modules. Document encoding is decoupled from concrete tasks. By plugging document plugins into task-specific models, we can handle multiple tasks such as question answering, fact verification, and slot filling.}
    % \vspace{-1.5em}
    \label{fig:example}
\end{figure}

For this question, we propose a novel framework based on PTMs to decouple document encoding from tasks, named \mymodel{}. Specifically, \mymodel{} incorporates plug-and-play modules to store document information and utilizes a PTM backbone to capture information from plugins for task reasoning.
As shown in Figure~\ref{fig:example}, documents are encoded into pluggable plugins once and for all before task adaptation. The semantics and knowledge of documents can be injected into task-specific models by plugging document plugins.
During task reasoning, the task-specific models can activate the information encoded in the document plugins to handle the input queries.
In this way, \mymodel{} can decouple the document encoding from downstream task reasoning and reduce the computation costs.

For representing documents as pluggable modules, there are two main challenges: (1)~Plugin learning: The document plugins must be effective for various downstream tasks, requiring them to contain sufficient semantics and knowledge. (2)~Plugin utilization: Once the document plugins are ready, it is important for task-specific models to capture relevant information from them effectively for task reasoning.

As for plugin learning, we adopt a self-supervised method, which requires document plugins to provide sufficient knowledge for the PTM to make predictions. Specifically, for each document, we first randomly select parts of sentences as a query and use the remaining sentences as context to learn plugins. Then, after encoding the context into plugins, the model is required to recover the masked recurring spans or generate the next sentences for the query based on the plugin knowledge. 

As for plugin utilization, we propose two strategies to utilize document plugins for downstream tasks: \textit{plugging during tuning} and \textit{plugging after tuning}\footnote{Here tuning refers to tuning PTMs for downstream tasks, including full-parameter fine-tuning and parameter-efficient tuning.}. 
For plugging during tuning, document plugins are utilized in both tuning and inference stages. Task data and document plugins are combined together to train task-specific models. 
For plugging after tuning, document plugins are only utilized in the inference stage to provide external knowledge. Document plugins are adopted as a post-processing way to inject knowledge into task-specific models without additional training.

To verify the effectiveness of our plug-and-play framework, we adopt Wikipedia as our document collection and conduct experiments on $8$ datasets of $4$ typical knowledge-intensive NLP tasks.
The results show that we can generate document plugins once and successfully adapt plugins to various downstream tasks. Compared to competitive baselines that encode documents and task-specific inputs simultaneously, our plugin-based method can save $69\%$ computational costs with comparable performance. Besides, utilizing document plugins works as an effective post-processing approach to introducing the knowledge of documents into task-specific models and achieving performance improvements without model training. 
We argue that with the current trend of increasing the model size of PTMs, decoupling document encoding from concrete tasks like \mymodel{} can be a promising direction that enables large PTMs to effectively and efficiently serve diverse downstream tasks.

\section{Related Work}

\subsection{Plug-and-Play Modules for PTMs}
Recent PTMs have shown to be effective in various downstream tasks~\cite{BERT,roberta,t5,GPT,gpt3,PTMs-survey,PaLM}. 
However, training and tuning large-scale PTMs for ever-increasing tasks is expensive in computation and storage. To address this issue, building plug-and-play modules with various capabilities for PTMs has received increasing attention recently. For instance, parameter-efficient tuning, which is also known as delta tuning, is proposed to perform task adaptation by fine-tuning only small amounts of parameters and keeping other parameters fixed~\cite{bitfit,adapter,prompt-tuning-powerofscale,prompt-survey,lora,Delta-tuning}.
The task-specific modules possess play-and-play characteristics and can effectively inject task ability into PTMs. 
Besides, some researchers explore combining pluggable modules with large-scale PTMs for efficient controllable text generation~\cite{PlugPLMControlGenration,DBLP:conf/emnlp/MadottoILDF20,DBLP:conf/emnlp/PascualEMCW21}, domain adaptation~\cite{DBLP:conf/naacl/ChronopoulouPD22,DBLP:conf/emnlp/PfeifferVGR20}, information retrieval~\cite{replug,yu2023augmentation}, knowledge injection~\cite{zhang2023plug}, model debias~\cite{DBLP:conf/emnlp/LauscherLG21}, and model integration~\cite{DBLP:journals/corr/abs-2305-08848,DBLP:conf/nips/AlayracDLMBHLMM22}. Owing to the powerful abilities of large-scale PTMs, these modules can effectively activate the model's capabilities with limited parameters.
Different from previous functional modules, we attempt to build document plugins to provide knowledge and context information for PTMs.

\subsection{Language Representation Learning}
Language representation learning is a fundamental NLP task ~\cite{representation-learning,BERT,GPT} that aims to effectively represent rich semantics distributed in text and benefit various downstream tasks.
Previous efforts attempt to map the language inputs into intermediate distributed features, such as word embeddings~\cite{word2vec,DBLP:conf/nips/KirosZSZUTF15,glove,ELMO}, sentence embeddings~\cite{DBLP:conf/emnlp/ConneauKSBB17,sentencebert,simcse}, and document embeddings~\cite{DBLP:journals/corr/DaiOL15,DBLP:conf/emnlp/WuYXXBCRW18}, which are further used as inputs of downstream task-specific models to generate the final task-specific document representations. Furthermore, some researchers make preliminary exploration to decouple document encoding from tasks by freezing the part of layers of document encoders~\cite{generalemb,embrecy}. But these works only achieve semi-decoupling of document encoding from tasks, and can only be used for the plugging during tuning setting.

Notably, many efforts have been devoted to exploring the effective architectures, such as sparse attention, of PTMs to encode long documents~\cite{longformer,bigbird,poolingformer,s4,efficientTransformer}. These works are parallel to ours, and we can adopt sparse-attention layers to further improve efficiency.

\section{Methodology}

In this section, we will first present the paradigm description and the overall framework of \mymodel{}
Then we introduce the self-supervised plugin learning method to make document plugins contain sufficient semantics and two strategies about how to utilize document modules.

\subsection{Plug-and-Play Document Modules}
In this paper, we focus on decoupling document encoding with specific tasks.
Different from encoding-task coupling methods which simultaneously encode the documents and task-specific queries, \mymodel{} aims to encode documents once and for all before task adaptation. 
Specifically, given a PTM backbone $\mathcal{M}$ and a document $d$, we first use the PTM to encode the document into a task-agnostic pluggable module, $\mathcal{D}$, i.e., a document plugin. Equipped with the document plugin, the PTM is injected into the corresponding document knowledge. Then we adopt task data to tune the PTM to obtain task-specific models.
During inference, we can quickly obtain predictions for an input query by inserting the relevant document plugin into the task-specific models, avoiding re-encoding the document.

\begin{figure*}[t]
    \centering
    \includegraphics[width=0.9\linewidth]{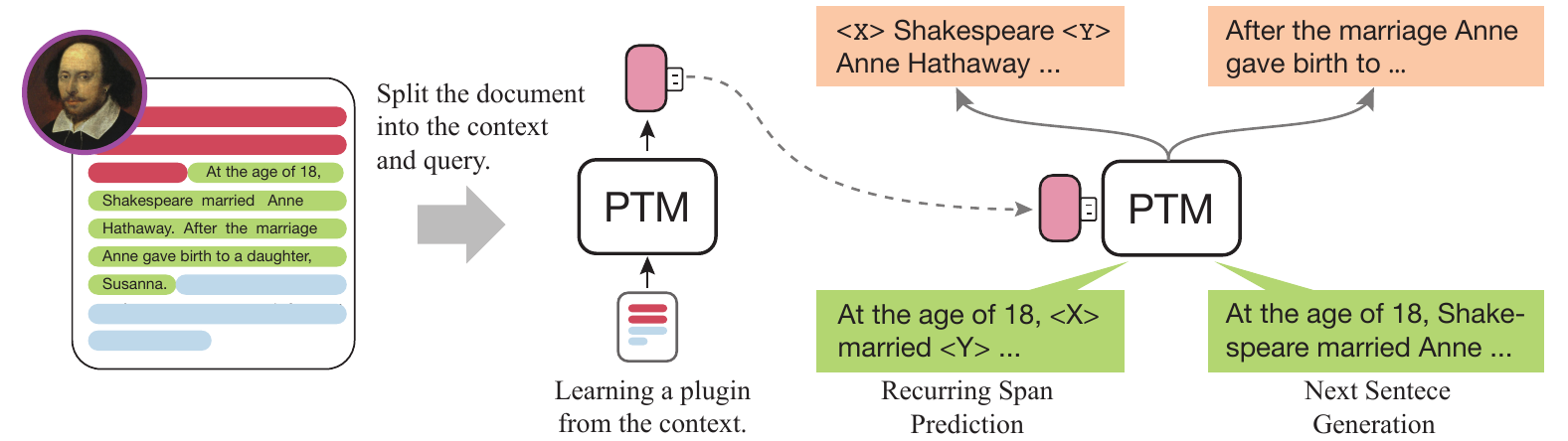}
    \caption{The illustration of \mymodel{} in plugin learning.}
    \vspace{-1em}
    \label{fig:main}
\end{figure*}

\subsection{Overall Framework}
As shown in Figure~\ref{fig:example}, we design \mymodel{}, which consists of three components: a PTM backbone, document plugins that provide document knowledge, 
and task-specific models derived from the PTM to handle specific tasks.
We will present these components below.

\textbf{PTM Backbone.}
PTMs have been proven effective in a wide range of downstream tasks, and raise a paradigm shift to solve multiple tasks with one unified model~\cite{foundation-model,gpt3,PaLM}. In view of this, we further explore the decoupling of document encoding and tasks, unifying document representations across tasks. \mymodel{} relies on a large-scale PTM, which can serve as a fundamental infrastructure to learn plugins from documents and as an initialization for task-specific models. Note that, for our framework, any PTM with large-scale parameters can be used as the backbone. Specifically, we adopt a widely-used sequence-to-sequence PTM, T5~\cite{t5}, in this paper.
As the pre-training objectives of the PTM do not involve the document plugins, we further conduct plugin learning for the PTM so that it can generate and utilize document plugins. The training tasks are introduced in the following sections.

\textbf{Document Plugin.} 
Document plugins store document knowledge and are obtained before utilizing these documents for specific tasks. Inspired by recent progress in model interpretation~\cite{LAMA,DBLP:journals/tacl/JiangXAN20,DBLP:conf/emnlp/RobertsRS20,knowledge-neuron,model-edit}, which claims that the parameters of PTMs store vast amounts of knowledge, we propose to encode the semantics and knowledge of documents into pluggable parameters. In this way, when the document plugin is inserted into the PTM, the PTM is empowered with the corresponding document knowledge.

Inspired by prefix-tuning~\cite{prefix-tuning}, we represent documents as prefix tokens for attention layers. When the document plugin is inserted into the backbone, we concatenate the corresponding prefix tokens with the hidden vectors of task-specific queries in attention layers to provide document knowledge.
Specifically, given a document $d$ with $L_d$ tokens, we first encode the document with the PTM to get the raw document representations $\bm{H}_d = \{\bm{h}_1, ..., \bm{h}_{L_d}\}$. 
Then, we adopt a mapping network to project the representation vectors into prefix tokens: $\bm{P}_d = \{\bm{p}_1, ..., \bm{p}_{L_d}\}$, where $\bm{p}_i = \bm{h}_i + \text{MLP}(\bm{h}_i)$. The prefix tokens are further inserted into the attention layers.
Let $\bm{H}_q = \{\bm{h}^q_1, ..., \bm{h}^q_{L_q}\}$ denote the hidden vectors of query in the attention layer. We calculate the attention output as follows:
\begin{equation}
\small
    \bm{H}_q^o = \text{Attn}(\bm{H}_q\bm{W}_q, \text{cat}(\bm{P}_d, \bm{H}_q)\bm{W}_k, \\ \text{cat}(\bm{P}_d, \bm{H}_q)\bm{W}_v),
\label{eq:attention}
\end{equation}
where $\bm{W}_q$, $\bm{W}_k$, and $\bm{W}_v$ are trainable parameters for the self-attention layer. Then $\bm{H}_q^o$ is fed into the feed-forward layer as the original Transformer~\cite{transformer} layer.

Different from encoding documents during task adaptation or inference, prefix tokens do not involve the computation of feed-forward layers.
Moreover, to better integrate the semantics of documents and queries for handling tasks, document plugins are only inserted in the near-top layers of the PTM backbone. Therefore, these document plugins in the form of prefix tokens only increase limited computational requirements, whereas \mymodel{} can achieve a significant computational speedup as a result.
Due to the high storage requirement of adding different prefix tokens to different attention layers, we share $\bm{P}_d$ for all attention layers.
Note that, we can also utilize other model structures, such as bias parameter~\cite{bitfit} and LoRA~\cite{lora}, to represent documents in \mymodel{}, which we leave for future work.

\textbf{Task-specific Models.}
Task-specific models are derived from the PTM backbone and tuned on the supervised task data to obtain task reasoning ability. During downstream tuning, we freeze the document plugins, and only the task-specific models and the mapping network of the document plugins are trainable so that the document plugins can be reused across different tasks.
We adopt two training methods for task-specific models, including vanilla full-parameter fine-tuning and parameter-efficient tuning (PET). Note that, deploying large-scale PTMs with full-parameter fine-tuning will lead to exacerbated computational and storage burdens for multi-task scenarios. Thus, it is worth exploring \mymodel{} with PET for efficient task adaption in real-world applications.

Both two training methods adopt task-specific objectives to optimize the parameters. Fine-tuning optimizes all parameters of the PTM backbone, while parameter-efficient tuning only optimizes parts of the parameters and keeps other parameters frozen.
Specifically, we adopt adapters for parameter-efficient tuning~\cite{adapter-ffn}. Given the hidden vector $\bm{h} \in \mathbb{R}^d$, where $d$ is the hidden size, the output of the adapter layer is calculated as:
\begin{equation}
\small
    \bm{h}_{out} = \bm{h} + \phi(\bm{h}\bm{W}_\text{down})\bm{W}_\text{up},
\end{equation}
where $\bm{W}_\text{down} \in \mathbb{R}^{d\times r}$, $\bm{W}_\text{up} \in \mathbb{R}^{r\times d}$, and $r \ll d$ refer to the bottleneck dimension.

\textbf{Computational Complexity.}
\mymodel{} encodes the documents before task adaptation and thus can reduce the computation costs. In this paragraph, we discuss the computational complexity of \mymodel{} in detail. Assume the lengths of the query and document are $L_q$ and $L_d$, respectively. For the traditional encoding-task coupling models, which simultaneously encode documents and queries, the computational complexity of the attention layer is $O((L_q + L_d)^2)$, and the computational complexity of the feed-forward layer is $O(L_q + L_d)$. For \mymodel{}, the document plugins are inserted into the attention layer, whose computational complexity is $O(L_q(L_q + L_d))$. And the document plugins do not involve the computation of the feed-forward layer, and thus its computational complexity is $O(L_q)$. In real-world applications, the documents usually are much longer than the queries. Therefore, \mymodel{} can achieve significant computational speedup compared with conventional encoding-task coupling models.

\subsection{Plugin Learning}
To enable the document plugins to contain sufficient document knowledge, we futher explore self-supervised plugin learning in this section.
As shown in Figure~\ref{fig:main}, we adopt two self-supervised tasks, recurring span prediction, and next sentence generation to augument the comphrension and generation ability of \mymodel{}. 
Both two tasks require document plugins to provide context information for the model to make the predictions. 
Let $d = \{s_1,...,s_n\}$ denote the input document with $n$ sentences. We perform plugin learning as:

\textbf{Recurring span prediction (RSP)}. Inspired by \citet{fewshotqa}, we utilize recurring spans to construct self-supervision signals.
Recurring spans occur multiple times in the documents, and usually contain important semantics for document understanding. Masking the recurring spans and requiring the PTM to recover them can help the PTM to capture document semantics. Specifically, we concatenate sentences randomly sampled from the document $d$ as query $q$, and treat the remaining sentences as context $c$. Then we generate the document plugin $\bm{P}_c$ based on $c$, and replace the recurring spans in $q$ as special mask tokens. The PTM is required to predict the masked spans in $q$ conditioned on $\bm{P}_c$. Different from the traditional masked language model task~\cite{BERT,t5}, which mainly focuses on local information around the masked spans, RSP usually requires the PTM to integrate global information from document plugins.

\textbf{Next sentence generation (NSG)}. To enable the document plugins to benefit generation tasks, we adopt NSG as a training task.
We first randomly sample three consecutive sentences $\{s_i, s_{i+1}, s_{i+2}\}$ from the document $d$. The remaining sentences are treated as the context $c = \{s_1, ..., s_{i-1}, s_{i+3}, ..., s_n\}$ to generate the document plugin $\bm{P}_c$. Then we regard $s_i$ as the query, and require the PTM to generate the following two sentences $\{s_{i+1}, s_{i+2}\}$ conditioned on $\bm{P}_c$.

These two tasks require the PTM to capture both local information from the queries and global information from the document plugins. Therefore, after plugin learning, the PTM is supposed to be able to build informative document plugins and serve as a good initialization for task-specific models to capture knowledge from document plugins.
Both two tasks are sequence-to-sequence tasks, and we adopt the negative likelihood as the training objectives for two tasks.
The model is trained in a multi-task fashion, and the final training loss is the sum of the loss of two tasks.
During plugin learning, the document plugins are calculated in real time for different documents. 
All parameters of the PTM are tuned for plugin learning. After that, the document plugins can be calculated and stored for further downstream task tuning and inference.

\subsection{Plugging Strategies}
To flexibly utilize the document plugins, we propose two plugging strategies:

\textit{Plugging during tuning.} 
In this setting, the document plugins are adopted in both the training and inference of task-specific models. 
Given an instance with the query and document as inputs, we first insert the corresponding document plugin, which is computed before fine-tuning, into the models. Then task-specific models are trained with task-specific objectives to capture relevant information from the document plugins.

\textit{Plugging after tuning.} 
In this setting, the document plugins are adopted only in the inference of task-specific models. Document plugins can provide external knowledge, and serve as a post-processing method to inject knowledge into task-specific models.
During inference, given an instance, we directly insert related document plugins into the task-specific models to achieve knowledge injection. This setting does not require additional training for existing task-specific models and can be used to flexibly inject textual knowledge.

\begin{table*}[t]
    \centering
    \small
    \begin{tabular}{l|cccccccccccc}
    \toprule
    \multirow{2}{*}{Models} 
     & FEVER & \multicolumn{2}{c}{NQ} & \multicolumn{2}{c}{TriviaQA} & \multicolumn{2}{c}{HotpotQA} & ELI5 & WoW & zsRE & T-Rex  & \multirow{2}{*}{Avg.}  \\
             & Acc. & EM & F1 & EM & F1 & EM & F1 & RL & F1 & Acc. & Acc. & \\
    \midrule
    & \multicolumn{12}{c}{Parameter-efficient Tuning} \\ \midrule
    ED2LM    & 83.13 & 38.34 & 46.04 & 53.84 & 62.05 & 19.84 & 28.63 
             & 11.24 & 15.24 & 31.15 & 46.34 & 37.39\\
    EmbRecy  & 84.59 & 37.42 & 45.43 & 53.02 & 61.05 & 18.98 & 27.70
             & 11.57 & 16.91 & 27.20 & 44.16 & 36.73 \\
    ED2LM$_f$$^\clubsuit$ & 81.81 & 35.62 & 44.18 & 52.01 & 59.82 & 19.07 & 27.81 & 11.01 & 15.20 & 27.09 & 44.78 & 35.82 \\
    EmbRecy$_f$$^\clubsuit$ & 84.59 & 32.13 & 40.17 & 47.59 & 55.37 & 18.18 & 26.79 & \textbf{11.92} & 16.65 & 20.76 & 41.22 & 34.13 \\
    \mymodel{}$^\clubsuit$ & \textbf{86.56} & \textbf{41.54} & \textbf{49.76} & \underline{57.29} & \textbf{65.43} & \textbf{23.04} & \textbf{32.51} 
             & 11.37 & \textbf{17.15} & \textbf{32.12} & \textbf{48.38} & \textbf{39.68}\\ 
    \quad w/o PT$^\clubsuit$ & \underline{86.33} & \underline{40.24} & \underline{47.72} & \textbf{57.67} & \underline{64.91} & \underline{22.04} & \underline{31.44} & \underline{11.67} & \underline{17.07} & \underline{30.64} & \underline{48.26} & \underline{39.24} \\ 
    \midrule
    \textcolor{gray}{UpperBound} & \textcolor{gray}{88.20} & \textcolor{gray}{42.60} & \textcolor{gray}{50.86} & \textcolor{gray}{61.77} & \textcolor{gray}{69.14} & \textcolor{gray}{23.84} & \textcolor{gray}{33.71} & \textcolor{gray}{11.80} & \textcolor{gray}{17.92} & \textcolor{gray}{33.65} & \textcolor{gray}{49.96} & \textcolor{gray}{41.22} \\ 

    \midrule \midrule

    & \multicolumn{12}{c}{Full-parameter Fine-tuning} \\ \midrule
    ED2LM  & 80.59 & 42.07 & 49.79 & 58.94 & 66.68 & 22.80 & 32.32 & 11.66 & 16.10 & \textbf{31.77} & 50.84 & 39.35 \\
    EmbRecy & 84.34 & \textbf{42.71} & \textbf{50.55} & \underline{59.31} & 66.67 & \textbf{23.57} & \textbf{33.46} & 12.01 & 17.30 & 30.10 & 50.12 & 39.93 \\
    ED2LM$_f$$^\clubsuit$ & 84.17 & 40.84 & 48.57 & 57.05 & 64.92 & 21.61 & 30.70 & 11.94 & 15.83 & 24.19 & 48.04 & 37.96 \\
    EmbRecy$_f$$^\clubsuit$ & 85.04 & 39.89 & 47.58 & 57.91 & 65.37 & 21.59 & 30.92 & 11.92 & 16.69 & 27.82 & 50.28 & 38.89 \\
    \mymodel{}$^\clubsuit$ & \textbf{86.34} & \underline{42.53} & \underline{50.42} & \textbf{59.46} & \textbf{67.07} & \underline{23.46} & \underline{33.07} & \textbf{12.30} & \textbf{17.61} & \underline{30.99} & \underline{52.22} & \textbf{40.61} \\
    \quad w/o PT$^\clubsuit$ & \underline{85.97} & 42.25 & 49.80 & 58.88 & 66.60 & 23.05 & 32.20 & \underline{12.16} & \underline{17.40} & 29.94 & \textbf{52.40} & \underline{40.26} \\
    \midrule
    \textcolor{gray}{UpperBound} & \textcolor{gray}{86.42} & \textcolor{gray}{45.03} & \textcolor{gray}{52.92} & \textcolor{gray}{62.50} & \textcolor{gray}{69.82} & \textcolor{gray}{24.54} & \textcolor{gray}{34.66} & \textcolor{gray}{12.33} & \textcolor{gray}{18.39} & \textcolor{gray}{32.60} & \textcolor{gray}{52.50} & \textcolor{gray}{41.79} \\
    \bottomrule
    \end{tabular}
    \caption{The main results of our proposed \mymodel{} and baselines for plugging during tuning. We boldface the best result and underline the second-best results for each dataset. The methods that can generate task-agnostic document representations are denoted with $^\clubsuit$.}
    % \vspace{-1.5em}
    \label{tab:main_results}
\end{table*}

\section{Experiments}

\subsection{Evaluation Settings}
\textbf{Datasets.}
We adopt widely-used Wikipedia articles as our document collection and select typical knowledge-intensive tasks for evaluation. 
We adopt a typical multi-task benchmark, KILT~\cite{KILT}, to evaluate our models. The tasks in KILT are grounded in the same snapshot of Wikipedia pages. In particular, we evaluate \mymodel{} on a fact verification dataset, FEVER~\cite{FEVER}, four question answering datasets, including Natural Questions (NQ)~\cite{NQ}, HotpotQA~\cite{hotpotqa}, TriviaQA~\cite{triviaqa}, ELI5~\cite{eli5}, a dialogue generation dataset, Wizard of Wikipedia (WoW)~\cite{WoW}, and two slot filling dataset, Zero Shot RE (zsRE)~\cite{zsRE}, T-REx~\cite{trex}. These tasks are diverse and require the model to exploit document knowledge fully. As shown in the paper of KILT, external document knowledge can not benefit the entity linking task.
Thus, we do not use them for evaluation in this paper.
Following \citet{KILT}, we use dense passage retrieval~\cite{dpr} to retrieve relevant documents from Wikipedia articles for each query. Please refer to Appendix for evaluation results of document retrieval.

\textbf{Metrics.}
Following previous work, we adopt accuracy for the fact verification task (FEVER) and slot filling tasks (zsRE, T-REx); exact match (EM) and F1 score for the extractive question answering tasks (NQ, HotpotQA, TriviaQA); ROUGE-L (RL) for the long abstractive question answering tasks (ELI5); F1 score for the dialogue generation task (WoW). Besides, to evaluate the overall performance, we calculate average scores for these tasks as an evaluation metric, in which EM scores are used for extractive question answering tasks.

\subsection{Training Details}
We utilize the widely used T5-large~\cite{t5}, as our PTM backbone. For the PET training method, we set the bottleneck dimension of adapters as $16$. % As mentioned before, document plugins are inserted in the near-top attention layers. 
We insert document plugins in the last $12$ layers.
We conduct plugin learning on a large-scale unsupervised corpus, C4~\cite{t5} for $36$k steps. We use Adam to optimize our models. Due to the high computational costs of full-parameter fine-tuning, in the following experiments, we adopt the PET method to train the models unless otherwise specified. We train models with a half-precision floating point on $8$ NVIDIA A100 GPUs, and the plugin learning process takes 18 hours.
Please refer to Appendix for more details.

\subsection{Baselines}
\textbf{Plugging during tuning.}
Here we compare \mymodel{} with several representative baselines, which encode the documents and queries with two different encoders. In this way, these models can reuse document representations across different queries, but they still need to generate different document representations for different tasks.
(1)~\textbf{ED2LM}~\cite{ed2lm} utilizes the encoder-decoder architecture to encode the queries and documents separately, and then the document can be pre-encoded before inference. In particular, the documents are inputted into the encoder, and queries are inputted into the decoder. 
(2)~\textbf{EmbRecy}~\cite{embrecy} proposes to reuse the intermediate activations of the documents to achieve speedup for fine-tuning and inference. EmbRecy caches an intermediate layer’s output as the document representation and the remaining near-top layers are tuned to fuse the information of documents and queries.
(3)~Besides, to meet the setting of decoupling document encoding from tasks, we freeze the document encoders of ED2LM and EmbRecy to make the document representations unified across different tasks. We denote the two task-agnostic methods as \textbf{ED2LM$_f$} and \textbf{EmbRecy$_f$}.
(4)~As \mymodel{} conducts further self-supervised training for \setting{}, we also present the results of \mymodel{} without plugin learning (\textbf{w/o PT}) to show the effectiveness of the architecture of \mymodel{}.
(5)~\textbf{UpperBound} follows the traditional settings, in which the queries and documents are concatenated together and fed into the model. The document representations generated by this baseline are query-specific. The model needs to encode a single document multiple times for different tasks and different queries, which is the upper bound of task-agnostic methods.

\begin{table}[t]
    \centering
    \small
    % \normalsize
    \begin{tabular}{l|ccc}
    \toprule
    \multirow{2}{*}{Models} & \multirow{2}{*}{Avg.} & FLOPs & Time \\ 
    & & G & ms  \\ \midrule
    ED2LM  & 37.39 & \textbf{114.9} & \textbf{60} \\
    EmbRecy & 35.54 & 197.5 & 142 \\
    \mymodel{} & \textbf{39.68} & \underline{139.3} & \underline{98} \\ \midrule
    UpperBound & 41.22 & 453.1 & 226 \\ \bottomrule
    \end{tabular}
    \caption{The average scores and computational costs of \mymodel{} and baseline models.}
    % \vspace{-1em}
    \label{tab:computation}
\end{table}

\textbf{Plugging after tuning.}
We attempt to inject unstructured textual knowledge into PTMs after downstream tuning. Existing methods mainly focus on enhancing PTMs with structural knowledge during pre-training or fine-tuning~\cite{ernie,kadapter,DBLP:conf/acl/BosselutRSMCC19}. These methods require retraining the task-specific models to achieve knowledge injection, which thus cannot be adopted in this setting. Therefore, we present the results of the following models: 
(1)~We adopt \textbf{T5}~\cite{t5} and \textbf{\mymodel{}} as the backbones, which are trained with only the queries as inputs and do not utilize external document knowledge in evaluation.
(2)~Based on the trained T5 and \mymodel{}, we adopt different post-processing methods to incorporate document knowledge. For T5, we directly concatenate the documents and queries as inputs for evaluation (\textbf{+Concat}). For \mymodel{}, we insert the document knowledge with document plugins (\textbf{+DPlug}).
The setting is challenging as there is a gap between the training and evaluation.

\subsection{Plugging during Tuning}
We present the comparison results between baseline models and \mymodel{} in Table~\ref{tab:main_results}. From this table, we can observe that: 
(1)~The baseline models which generate task-agnostic document representations perform worse than the corresponding models which generate task-specific representations. It indicates that decoupling document representation from concrete tasks is challenging and existing methods cannot achieve satisfactory performance.
(2)~Compared with the task-agnostic baseline models (ED2LM$_f$ and EmbRecy$_f$), \mymodel{} can achieve significant performance improvements across different tasks. Besides, compared with ED2LM and EmbRecy, \mymodel{} can also achieve superior results on many datasets, especially for parameter-efficient tuning. In addition, ED2LM and EmbRecy need to generate document representations for different tasks separately. Thus they require more storage than \mymodel{}. In contrast, \mymodel{} can generate informative unified representations with fewer storage requirements and achieve superior results across different tasks.
(3)~Compared with the traditional encoding-task coupling model (UpperBound), sharing document representation across different tasks in \mymodel{} only leads to a limited performance drop ($39.68$ vs. $41.22$, and $40.61$ vs. $41.79$ on average). And as \mymodel{} does not need to encode documents during downstream tuning and inference, \mymodel{} enables significant computational acceleration. The results suggest that \mymodel{} can effectively capture document semantics and inject them into the PTM to provide knowledge. 
(4)~Even \mymodel{} without further plugin learning can outperform the baselines on several datasets. It proves that \mymodel{} benefits from both the self-supervised tasks and the model architecture. Besides, it also indicates that the contextualized document representations generated by the original PTM (\mymodel{} w/o PT) are powerful if we utilize them correctly.

\begin{table}[t]
    \centering
    \small
    \resizebox{\linewidth}{!}{
    \begin{tabular}{l|rrrrrr}
    \toprule
    \multirow{2}{*}{Models} 
     & FEVER & \multicolumn{2}{c}{NQ} & WoW & zsRE \\
             & Acc. & EM & F1 & F1 & Acc. \\
    \midrule
    T5 & 79.10 & 11.35 & 17.11 & 16.59  & 2.52 \\
    +Concat   & 76.84 & 14.45 & 22.16 & 14.26 & 19.17 \\
    $\Delta$ & -2.26 & +3.1 & +5.05 & -2.33 & +16.65 \\ \midrule
    \mymodel{} & 79.56 & 11.17 & 16.39 & 16.58 & 2.23 \\
    +DPlug & 82.54 & 23.01 & 32.68 & 15.28 & 21.13 \\
    $\Delta$ & \textbf{+2.98} & \textbf{+11.84} & \textbf{+16.29} & -1.03 & \textbf{+18.90}
    \\ \bottomrule
    \end{tabular}}
    \caption{The main results of our proposed \mymodel{} and baselines for plugging after tuning.}
    % \vspace{-1.5em}
    \label{tab:plug-after}
\end{table}

\textbf{Computational Cost.}
We compare the computational cost of \mymodel{} and baseline models. Here, we present the floating point operations (FLOPs) and calculation time required to process one data in inference for each method. We assume that the document, query, and answer contain $512$, $48$, and $32$ tokens, respectively. 
The results are shown in Table~\ref{tab:computation}. From the results, we can observe that: 
(1) The methods for task-agnostic representation require much less computational cost than encoding-task coupling methods. Especially, our method \mymodel{} can achieve $3.25 \times$ speed up ($139.3$ GFLOPs vs. $453.1$ GFLOPs). 
(2) The methods for task-agnostic representation generally are inferior to encoding-task coupling methods. \mymodel{} can achieve better average scores than other baselines and preserve low computational costs. 
(3) Both task-agnostic and query-agnostic models need to pre-encode and store document representations before downstream tuning for inference speed up. However, models generating query-agnostic and task-specific representations require separate document representations for each task. In contrast, our \mymodel{} generates task-agnostic representations for all tasks, resulting in better results and lower storage requirements.

\begin{table}[t]
    \centering
    \small
    \begin{tabular}{l|ccccc}
    \toprule 
    Datasets & FEVER & \multicolumn{2}{c}{NQ} & WoW & zsRE \\
    & Acc. & EM & F1 & F1 & Acc. \\
    \midrule
    \mymodel{} & \textbf{86.56} & \textbf{41.54} & \textbf{49.76} & 17.15 & \textbf{32.12} \\
    ~~w/ RSP   & 86.17 & 41.23 & 49.21 & 16.98 & 31.66 \\
    ~~w/ NSG   & 86.03 & 40.80 & 49.06 & \textbf{17.62} & 28.92 \\
    ~~w/o PT   & 86.33 & 40.24 & 47.72 & 17.07 & 30.64 \\
    \bottomrule
    \end{tabular}
    \caption{The results of ablation study.}
    % \vspace{-1.5em}
    \label{tab:ablation}
\end{table}

\subsection{Plugging after Tuning}
The comparison results are shown in Table~\ref{tab:plug-after}. From the results, we can observe that:
(1)~Both T5 and \mymodel{} cannot achieve consistent improvement from post-processing knowledge injection on these tasks. It proves that plugging after tuning is a challenging setting as there is a gap between training and evaluation.
(2)~\mymodel{} can achieve significant improvement on FEVER, NQ, and zsRE, which further indicates the effectiveness of \mymodel{}. However, \mymodel{} cannot achieve improvement on WoW. As the ability to acquire knowledge from the document plugins is obtained from plugin learning, further downstream task tuning may lead the models to forget the ability. Thus, even \mymodel{} can not achieve consistent improvement. 
(3)~Without document knowledge, \mymodel{} and T5 achieve comparable results. It indicates that the plugin learning process does not improve the fundamental ability of PTMs. The improvement achieved by \mymodel{} in both plugging during/after tuning settings comes from the effective plug-and-play framework.

\subsection{Ablation Study}
In this section, we conduct an ablation study to verify the effectiveness of our proposed plugin learning tasks. We show the results of the models, which are trained with only recurring span prediction task (\textbf{w/ RSP}), with only next sentence generation task (\textbf{w/ NSG}), or without plugin learning (\textbf{w/o PT}). We evaluate the models on four datasets for the plugging during tuning setting.

The results are shown in Table~\ref{tab:ablation}. We can find that 
(1)~\mymodel{} without plugin learning leads to a significant performance drop, which further indicates that the proposed training tasks can help the PTM to effectively encode the document knowledge into plugins.
(2)~Two tasks can cooperate with each other to improve the model performance. Though training \mymodel{} with only one task will lead to performance deterioration on some tasks, training with two tasks can achieve consistent improvement over the model without plugin learning.
(3)~When \mymodel{} is trained with only NSG, the model can achieve superior results for WoW. But the task harms the performance for FEVER and zsRE. This is because NSG requires the model to generate long sentences, which is similar to WoW, while FEVER and zsRE only require short outputs. In contrast, training with only RSP will also lead to a performance drop for WoW.
It indicates that diverse plugin learning tasks are important for expressive document plugins.

\begin{figure}[t]
    \centering
    \hspace{-0.1\linewidth}\includegraphics[width=0.9\linewidth]{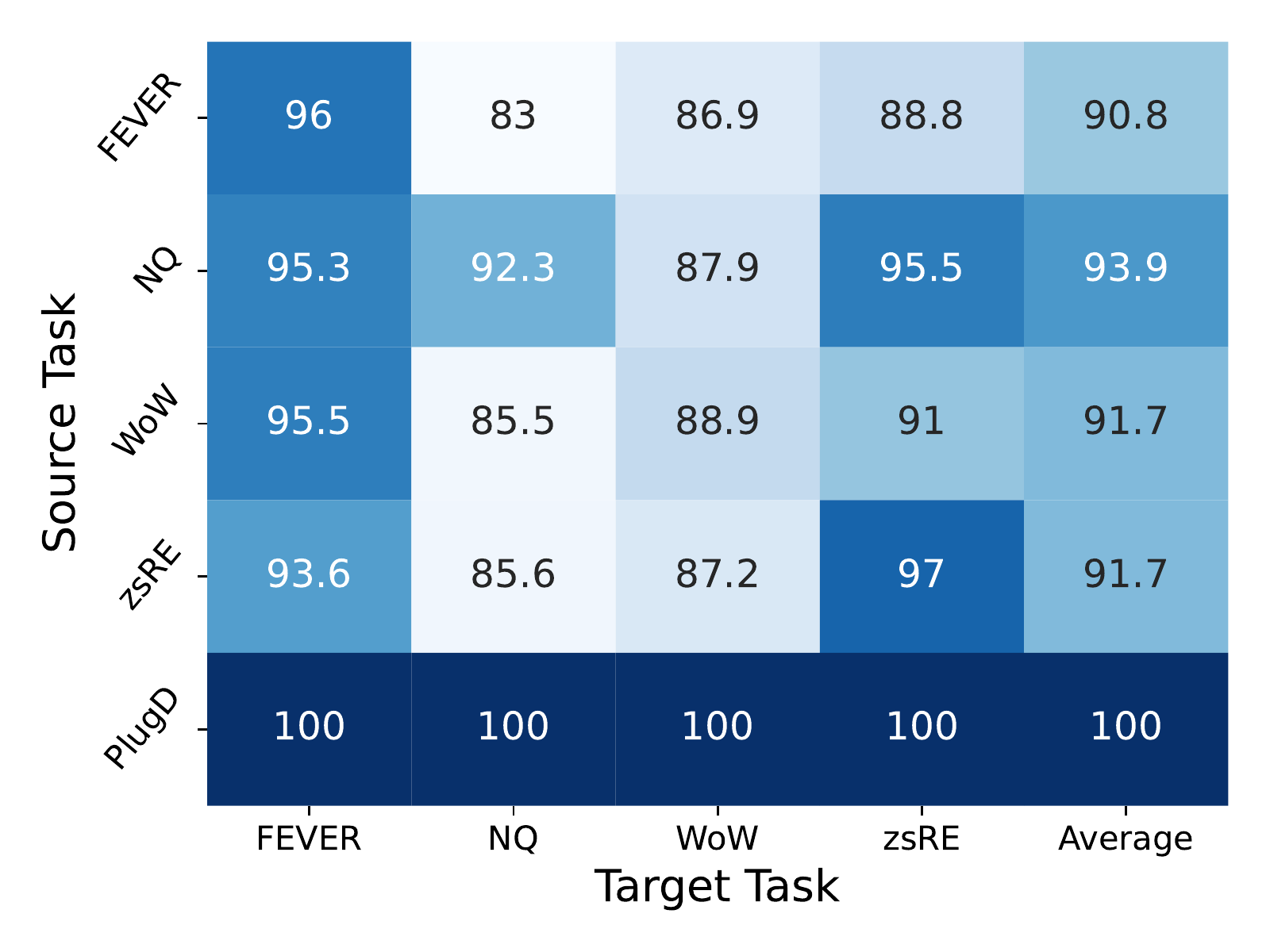}
    \caption{Relative transfer performance (transfer performance / \mymodel{}'s performance)(\%).}
    % \vspace{-1.5em}
    \label{fig:transfer}
\end{figure}

\subsection{Transferability Analysis}
In this section, we want to explore the effectiveness of supervised tasks on document representation transferability.
Here we present the results of ED2LM, which can outperform other baselines.
Specifically, we train the task-specific document encoder on a source task, and then reuse the encoder on other target tasks to continually train the rest of the model.
The results are shown in Figure~\ref{fig:transfer}.

From the results, we can observe that 1) The non-diagonal values of the matrix are consistently smaller than the diagonal values. It suggests that training the document encoder with existing supervised tasks can hardly benefit other target tasks. \mymodel{} trained with two self-supervised objectives can provide transferable document representation and achieve superior results. 2) The encoders trained on the NQ dataset can outperform encoders trained on other tasks. It indicates that training with challenging tasks may lead to better performance.

\section{Conclusion}
In this paper, we explore a new paradigm, which aims to represent documents as pluggable modules for PTMs. In this setting, we can get rid of encoding the same document multiple times for different tasks. The extensive experiments prove that our proposed \mymodel{} can significantly reduce the computational cost and effectively inject document knowledge into PTMs to improve performance. In the future, we will explore more effective plugin learning tasks and further attempt to represent knowledge graphs, and figures as plugins to provide knowledge for PTMs. 

\section*{Limitations}
We discuss the limitations of \mymodel{} in this section:
(1) We explore decoupling document encoding from concrete tasks in this paper, and propose to represent documents as pluggable modules before task adaptation. Therefore, \mymodel{} has a higher storage requirement than conventional encoding-coupling methods. We encourage 
(2) In the experiments, we adopt T5 as our PTM backbone. Actually, the proposed framework can also be applied to more pre-trained models with various model architectures. Besides, recent trends show that larger models tend to build more expressive text representations. It is worth exploring \mymodel{} with larger PTMs with billions of parameters to learn informative document plugins.
(3) In this paper, we adopt an external retriever to retrieve relevant documents for each input query. Recent progress in retrieval-augmented language models shows that training the PTMs with an end-to-end textual knowledge retriever can promote downstream performance. We believe document plugins can also serve as the external knowledge base and enhancing \mymodel{} with end-to-end retrieval is a promising direction. 

\section*{Acknowledgement}
This work is supported by the National Key R\&D Program of China (No. 2020AAA0106502), National Natural Science Foundation of China (No. 62236004).

\paragraph{Author Contributions} 
Chaojun Xiao and Chi-Min Chan wrote the code and conducted the experiments. 
Chaojun Xiao, Zhengyan Zhang, and Xu Han wrote the initial draft. 
Yankai Lin, Zhiyuan Liu, and Xiangyang Li significantly edited and improved the paper. 
Zhonghua Li, Zhao Cao, and Maosong Sun provided valuable advice to the research.

\bibliography{anthology,custom}
\bibliographystyle{acl_natbib}

\newpage
\appendix

\section{Appendix}

\subsection{Discussion}
% 主要讨论一下potential direction
% 1. 使用统一的模型来完成各种任务逐渐是一个非常重要的方向（ Instruct tuning），我们这个也是一种路径
% 2. 
In this paper, we propose to decouple document encoding from concrete tasks, achieving encoding documents once and for all across different tasks. In this section, we further discuss the potential of \mymodel{}.

\textit{Unified Model across Multiple Tasks.} Recently, with the rapid progress of large-scale PTMs, handling multiple tasks with a unified PTM has received rapidly increasing attention. For example, many researchers explore instruction tuning~\cite{DBLP:conf/iclr/SanhWRBSACSRDBX22,DBLP:conf/iclr/WeiBZGYLDDL22} to enable a unified PTM to perform multiple tasks with natural language description. We attempt to extend the paradigm to document representation, enhancing the unified PTM with unified document representation across multiple tasks. 
In this way, we can provide the PTM with various external knowledge flexibly and efficiently, avoiding encoding documents multiple times for different tasks and user input queries. 

\textit{Heterogeneous Knowledge Base.} Enhancing large-scale PTMs with various knowledge is an important topic for natural language processing. Many researchers attempt to incorporate knowledge graphs~\cite{ernie,kadapter}, linguistic knowledge~\cite{DBLP:conf/emnlp/Zhou0ZZ20} into PTMs. We argue that \mymodel{} provides a new way for knowledge injection. We can encode various knowledge, such as images, and knowledge graphs, into the plugins of PTMs. In this way, we can build a heterogeneous plugin knowledge base for PTMs to improve downstream performance.

\textit{Continual Learning.} Previous researches show that PTMs can implicitly encode knowledge in the model parameters~\cite{LAMA,DBLP:journals/tacl/JiangXAN20,DBLP:conf/emnlp/RobertsRS20,knowledge-neuron,model-edit}, which is not editable for continual updates. PlugD provides a new way for the continual learning of PTMs. We can insert and update new knowledge for PTMs by continually learning and updating new document plugins, which will be further utilized to provide knowledge to PTMs.

\begin{table*}[t]
    \centering
    \small
    \begin{tabular}{l|rrrrrrrr}
    \toprule 
    Datasets & FEVER & NQ & TriviaQA & HotpotQA & ELI5 & WoW & zsRE & TRex \\
    \midrule
    R-Precision   & 56.29 & 55.97 & 46.76 & 25.58 & 16.46 & 27.37 & 14.72 & 43.74 \\
    % P@1     & 53.59 & 55.97 & 46.76 &  3.82 & 16.46 & 27.37 & 14.72 & 43.74 \\
    Precision@3     & 24.01 & 27.71 & 24.82 &  2.82 & 10.62 & 15.04 &  6.87 & 18.40 \\
    Recall@3	    & 70.25 & 59.25 & 51.64 &  8.46 & 23.90 & 45.12 & 19.20 & 55.21 \\
    % P@5     & 15.52 & 19.27 & 17.32 &  2.33 &  9.17 & 10.60 &  4.69 & 11.75 \\
    % R@5     & 75.64 & 66.26 & 58.84 & 11.63 & 33.66 & 52.98 & 21.75 & 58.75\\
    \bottomrule
    \end{tabular}
    \caption{The results of document retrieval for each dataset.}
    \label{tab:layers}
\end{table*}

\subsection{Document Retrieval}
In this paper, we adopt dense passage retrieval, DPR~\cite{dpr}, to retrieve relevant documents for each input query. Following \citet{KILT}, we adopt R-Precision, Precision@k and Recall@k, as the evaluation metrics. We adopt the evaluation scripts provided by the KILT paper. Please refer to the original KILT paper for the details of the metrics. From the results, we can see that the retrieval performance is not satisfactory for some datasets, which may bring the noise in the downstream tasks. And we encourage the community to develop retrievers, which can achieve satisfactory performance across different tasks.

% \section{Hyper-parameter Analysis}
\subsection{Impacts of Insertion Layers}
\begin{table}[h]
    \centering
    \small
    \begin{tabular}{l|ccccc}
    \toprule 
    Datasets & FEVER & \multicolumn{2}{c}{NQ} & WoW & zsRE \\
    & Acc. & EM & F1 & F1 & Acc. \\
    \midrule
    \mymodel{} (6)  & 85.22 & 39.78 & 48.12 & 17.15 & 28.44 \\
    \mymodel{} (12) & 86.33 & 40.24 & 47.72 & 17.07 & 30.64 \\
    \mymodel{} (24) & 86.64 & 40.52 & 48.77 & 16.86 & 29.14 \\
    \bottomrule
    \end{tabular}
    \caption{The results of \mymodel{} with different number of insertion layers. Here \mymodel{} ($n$) indicates that the document plugins are inserted into the top-$n$ layers.}
    \label{tab:layers}
\end{table}

\mymodel{} inserts the document plugins into the self-attention layers to provide document knowledge. As the pre-trained models tend to capture linguistic features in the bottom layers and capture the task-specific information in the top layers~\cite{DBLP:journals/tacl/RogersKR20}. Therefore, to reduce computational costs, we only insert the document plugins in the top layers. In this section, we explore the impact of insertion layers of document plugins. We present the results of \mymodel{} with document plugins inserted in the last $6$ layers, $12$ layers, and all 24 layers. Here, we do not conduct plugin learning for \mymodel{} to speed up experiments.

The results are shown in Table~\ref{tab:layers}. From the results, we can see that: (1) With the increasing of insertion layers, the performance on FEVER and NQ improves. But \mymodel{} with document plugins in all layers can not outperform the \mymodel{} with document plugins in the top layers on WoW and zsRE. That is because the fact verification and question answering tasks require the models to select useful information via both lexical matching and semantic matching. In contrast, the dialogue generation and slot filling tasks rely on document semantics to provide knowledge, and inserting the document plugins in the bottom layers can not benefit the performance.
(2) The three models can obtain similar performance on these tasks. Therefore, in order to reduce the computational costs and maintain the performance, we only insert document plugins in the top $12$ layers for other experiments.

\subsection{Impacts of Plugin Sharing across Layers}
As mentioned in previous sections, \mymodel{} inserts the same prefix tokens for different attention layers to save the storage. In this section, we study the impacts of sharing plugins across different layers. To this end, we attempt to insert different prefix tokens for different layers. Specifically, we encode the document $d$ to obtain the raw hidden state $\bm{H}_d^l$ from the $l$-th layer, and then adopt the mapping network tailored to the $l$-th layer to map the hidden state into the prefix tokens. The prefix tokens are then inserted into the $l$-th layer for query encoding. Similar to \mymodel{}, we insert the representations into the top $12$ layers for this model. We term the model as All-Hidden.

The comparison results are shown in Table~\ref{tab:sharing}. From the results, we can observe that All-Hidden can achieve superior results on three datasets, including FEVER, WoW, and zsRE. But All-Hidden requires $12\times$ storage than \mymodel{}, which is impractical for large-scale document collections. And \mymodel{} can achieve comparable performance to All-Hidden. Therefore, to reduce the storage requirement, we choose to share the document plugins across different attention layers.

\subsection{Experimental Details}

\paragraph{Model Implementation.} The mapping network of document plugins is used to map the raw document representations into the document plugins for different tasks. Given a hidden vector, $\bm{h_i}$, we calculate the corresponding prefix token as $\bm{p_i} = \bm{h_i} + \bm{W_m^2}\text{ReLU}(\bm{W_m^1h_i})$, where $\bm{h_i} \in \mathbb{R}^d$, $\bm{W_m^1} \in \mathbb{R}^{d\times2d}$, $\bm{W_m^2} \in \mathbb{R}^{2d\times d}$, and $d \in \mathbb{R}$ is the hidden size. 

As for the parameter-efficient tuning method, we adopt adapter layers to tune the model. We add the adapters after the layernorm operation of feed-forward layers. The parameters of adapters are randomly initialized following a zero-mean Gaussian distribution with standard deviation as $10^{-2}$.

\paragraph{Plugin Learning.} For the recurring span prediction task, we first identify spans that occur multiple times from the documents. Then we filter out the stopwords and personal pronouns, and keep the longest $15$ spans as the recurring spans for further masking. Then we randomly sample $5$ sentences, which contain the recurring spans from the document as the query. For the next sentence generation task, we randomly sample three consecutive sentences from the documents, where the first sentence is treated as the query, and the last two sentences are treated as the answers. The model is trained in a multi-task fashion, and $70\%$ documents are used for recurring span prediction, and $30\%$ documents are used for next sentence generation. The maximal length for queries and answers are set as $196$ and $128$, respectively. We set the learning rate as $2\times10^{-5}$ and batch size as $256$. 

\begin{table}[t]
    \centering
    \small
    \begin{tabular}{l|ccccc}
    \toprule 
    Datasets & FEVER & \multicolumn{2}{c}{NQ} & WoW & zsRE \\
    & Acc. & EM & F1 & F1 & Acc. \\
    \midrule
    \mymodel{}  & 85.22 & 39.78 & 48.12 & 17.15 & 28.44 \\
    All-Hidden  & 86.73 & 39.46 & 47.71 & 17.28 & 32.20 \\
    \bottomrule
    \end{tabular}
    \caption{The comparison results of \mymodel{} and All-Hidden that does not share plugins across layers.}
    \label{tab:sharing}
\end{table}

\paragraph{Downstream Task Tuning.}
For downstream tasks, we set the training batch size as $64$. The learning rate is selected from $\{10^{-4}, 5\times10^{-4}, 10^{-3}\}$ for PET. And as full-parameter fine-tuning require amounts of computation, we do not conduct grid search for this setting. We set the learning rate for full-parameter fine-tuning as $2\times10^{-5}$. For fact verification, we take the claims as the input queries and take the logits of ``yes" and ``no" for classification. For other tasks, we treat them as text-to-text generation problems, and during the inference, we adopt the greedy strategy for decoding. The evaluation scripts are written by our own, and will be released with the paper.

\end{document}